\documentclass{article}
\usepackage{indentfirst} 
\usepackage{amsmath}
\usepackage{arxiv}

\usepackage[utf8]{inputenc} 
\usepackage[T1]{fontenc}    
\usepackage{hyperref}       
\usepackage{url}            
\usepackage{booktabs}       
\usepackage{amsfonts}       
\usepackage{nicefrac}       
\usepackage{microtype}      
\usepackage{lipsum}
\usepackage{graphicx}
\graphicspath{ {./images/} }
\title{\textbf{Cross-modal semantic segmentation for indoor environmental perception using single-chip millimeter-wave radar raw data} }

\author{
 Hairuo Hu \\
  School of Chemical Engineering \\
  Dalian University of Technology \\
  Dalian 116024, China \\
  \texttt{huhairuobiu@mail.dlut.edu.cn} \\
   \And
 Haiyong Cong \\
  School of Chemical Engineering \\
  Dalian University of Technology \\
  Dalian 116024, China \\
  \texttt{hycong@dlut.edu.cn} \\
  \And
 Zhuyu Shao \\
  School of Chemical Engineering \\
  Dalian University of Technology \\
  Dalian 116024, China \\
  \texttt{shaozhuyu@mail.dlut.edu.cn} \\
  \And
 Yubo Bi \\
  School of Chemical Engineering \\
  Dalian University of Technology \\
  Dalian 116024, China \\
  \texttt{byb@dlut.edu.cn} \\
  \And
 Jinghao Liu \\
  School of Chemical Engineering \\
  Dalian University of Technology \\
  Dalian 116024, China \\
  \texttt{liujinghao@dlut.edu.cn} \\
}

\begin{document}
\maketitle
\begin{abstract}
In the context of firefighting and rescue operations, a cross-modal semantic segmentation model based on a single-chip millimeter-wave (mmWave) radar for indoor environmental perception is proposed and discussed. To efficiently obtain high-quality labels, an automatic label generation method utilizing LiDAR point clouds and occupancy grid maps is introduced. The proposed segmentation model is based on U-Net. A spatial attention module is incorporated, which enhanced the performance of the mode. The results demonstrate that cross-modal semantic segmentation provides a more intuitive and accurate representation of indoor environments. Unlike traditional methods, the model's segmentation performance is minimally affected by azimuth. Although performance declines with increasing distance, this can be mitigated by a well-designed model. Additionally, it was found that using raw ADC data as input is ineffective; compared to RA tensors, RD tensors are more suitable for the proposed model.
\end{abstract}
\section{Introduction}
Since the 21st century, while society and the economy have developed rapidly, the number of fires has also increased annually \cite{ref1}. Statistics show that most firefighter casualties occur during fire rescues \cite{ref2}. In fire scenes, dim environments and heavy smoke severely limit the environmental perception of firefighters, negatively impacting rescue efficiency. Impaired perception can lead to disorientation and falls, causing injuries or fatalities, especially in complex indoor environments. Additionally, the air respirators for firefighters typically last only 30 minutes \cite{ref3}, after which they face the risk of asphyxiation or poisoning, making rescue efficiency even more critical. Therefore, enhancing the environmental perception in fire scenes is essential to reducing casualties and improving rescue efficiency. 

Indoor fire scene perception aims to use sensors to efficiently and accurately detect and visualize the environment, providing vital information for personnel and laying a foundation for post-processing applications. Currently, cameras \cite{ref4} and LiDAR \cite{ref5}.are the primary tools for indoor environment perception. However, in smoke-filled and dim environments, cameras fail to properly expose and capture images, and LiDAR suffers from scattering in smoke \cite{ref6}, making both unsuitable for fire scenes.. In recent years, Frequency Modulated Continuous Wave (FMCW) mmWave radar emerged as a promising solution for environmental sensing in autonomous driving \cite{ref7}, advanced driver-assistance systems (ADAS) \cite{ref8}, and automatic emergency braking (AEB) \cite{ref9}. Besides its low cost, its robustness in harsh conditions like fog and smoke \cite{ref10} makes it particularly suitable for fire scene perception. 

Essentially, indoor environmental perception forms the basis for mapping \cite{ref11,ref12}, localization \cite{ref13,ref14}, trajectory tracking \cite{ref15,ref16}, and navigation \cite{ref17}. For fire scenes, where traditional sensors fail, mmWave radar must independently perform environment perception. However, mmWave radar is limited by hardware constraints and signal processing methods The point clouds generated by mmWave radar are significantly inferior to those of LiDAR, not only being sparse and unstable \cite{ref18} but also prone to interference from multipath effects \cite{ref19}, particularly with single-chip mmWave radar. These limitations greatly hinder its application in indoor environment perception.

Various radar data processing methods have been proposed to improve point cloud accuracy. Constant False Alarm Rate (CFAR) detectors are widely used in point cloud generation, designed to detect target signals amid noise \cite{ref20}. However, CFAR algorithms, such as cell averaging (CA) CFAR \cite{ref21}, greatest of (GO) CFAR \cite{ref22}, ordered statistic (OS) CFAR \cite{ref23}, etc. are prone to false alarms and missed detections. Weak signals may contain rich environmental information, while strong signals could be cluttered. In complex indoor environments with multiple reflections, the detection performance of CFAR degrades significantly, and it cannot effectively eliminate multipath effects.

Direction of Arrival (DOA) estimation is another critical step in point cloud generation. Super-resolution algorithms like multiple signal classifier (MUSIC) \cite{ref24}, iterative adaptive approach (IAA) \cite{ref25}, etc. offer better accuracy than traditional Fast Fourier Transform (FFT)-based azimuth estimation methods, but they have high computational complexity and limited super-resolution capabilities. Recent advances in machine learning have introduced learning-based methods for DOA estimation and radar detection. Brodeski et al. \cite{ref26} a used an RCNN model for DOA estimation of single targets, and Cheng et al. \cite{ref18} developed RPDNet, a radar detector based on Range-Doppler Maps (RDM), which improves point cloud accuracy and density while mitigating multipath effects. However, the precision of point clouds still lags behind LiDAR. 

Semantic segmentation of navigable space, instead of point cloud-based obstacle detection, is another promising approach. Orr et al. \cite{ref27} proposed a cross-modal semantic segmentation model based on deep neural networks (DNNs). This model takes raw RD tensor as input and directly performs navigable space segmentation in the overlapping field of view (FoV) with camera images, demonstrating the potential of DNNs to interpret abstract mmWave radar signals and selectively filter them. Jin et al. expanded this work by adding multi-class functionality \cite{ref28} and introducing a transformer structure \cite{ref29,ref30}. While these efforts were conducted in outdoor driving scenarios, this idea also holds great potential for indoor environmental perception.

To achieve indoor fire scene perception, common indoor environment perception must first be realized, which is the focus of this paper. Point clouds are not the only solution for rescue operations. Direct segmentation of navigable space offers more intuitive guidance for personnel, eliminating the need for sparse point cloud-based obstacle detection and path planning. 

However, applying cross-modal semantic segmentation to indoor environment perception tasks faces several key obstacles. Previous studies primarily used high-resolution automotive radars, while the potential of the more affordable single-chip mmWave radar remains unexplored. Compared to high-resolution radars, single-chip radars have fewer antennas, providing less environmental data. Moreover, indoor environments are more complex, posing significant challenges for cross-modal models. The design of segmentation tasks and labels is crucial, as it directly affects model performance. This paper addresses these challenges and contributes as follows: 

 \begin{itemize}
     \item The feasibility of applying a cross-modal semantic segmentation model based on single-chip mmWave radar for indoor environment perception has been validated. A U-Net-based semantic segmentation model was proposed, demonstrating excellent performance on this task.
 \end{itemize}

 \begin{itemize}
     \item To reduce labor and time costs, standardize the labeling process, and improve label quality, an efficient automatic labeling method based on LiDAR point clouds and occupancy grid maps was introduced and implemented.
 \end{itemize}

 \begin{itemize}
     \item The point cloud-based environmental perception method and the proposed segmentation-based method were intuitively compared and evaluated. The characteristics of the proposed model and method for the segmentation task were thoroughly discussed from the perspectives of distance, azimuth, and scene complexity. A detailed analysis of their respective advantages and disadvantages was provided, along with suggestions for future optimization. Furthermore, the model's adaptability to different types of input data was also explored.
 \end{itemize}

It is important to note that this paper focuses on the feasibility of using segmentation models for indoor perception, with factors related to smoke left for future exploration. The remainder of this paper is structured as follows: Section 2 introduces the dataset and LiDAR-based labeling method. Section 3 describes the semantic segmentation model, and Section 4 outlines the training details. Section 5 compares the traditional point cloud method with the proposed method, evaluating model performance and discussing characteristics and ablation experiments. Finally, Section 6 concludes the paper.

\section{Data Preparation}
\subsection{ColoRadar Dataset}
All data used in this work comes from the ColoRadar dataset, a high-quality dataset published by Kramer et al. \cite{ref31}. It provides over two hours of diverse 3D scenes, both indoor and outdoor, totaling 52 sequences. The indoor scenes are categorized into four types: laboratories, indoor corridors, indoor open spaces and mining tunnels (see Fig. 1), each with distinctly different spatial structural characteristics. To prevent model bias towards any specific indoor scene, 5460 frames were selected from each category resulting in a total of 21840 frames. Of these, 20\% were reserved for the test set.

The ColoRadar dataset was collected using the 16-line Ouster OS1 LiDAR, single-chip FMCW mmWave radar AWR1843BOOST, and the 4-chip cascaded mmWave radar MMWCAS-RF-EVM. Several types of data were provided:  LiDAR point cloud data,  ADC raw data from the single-chip mmWave radar,  processed point clouds and heatmaps,  ADC raw data and heatmaps from the 4-chip cascaded mmWave radar. Additionally, IMU information and ground truth poses are provided, with all data accurately timestamped for precise synchronization.

In this study, data from the single-chip mmWave radar is used. This radar consists of three transmit antennas (Tx) and four receive antennas (Rx), forming 12 virtual channels through a time-division multiplexing multi-input single-output (TDM-MIMO) strategy. Under the configured data collection parameters, the radar system has a maximum detection range of 12 m and a maximum Doppler velocity of 2.56 m/s. The theoretical resolution for range, Doppler velocity, horizontal angle, and vertical angle are 0.125 m, 0.04 m/s, 11.3°, and 45° respectively. For more detailed information on the equipment models and parameters, refer to reference \cite{ref31}.
\begin{figure} 
    \centering
    \includegraphics{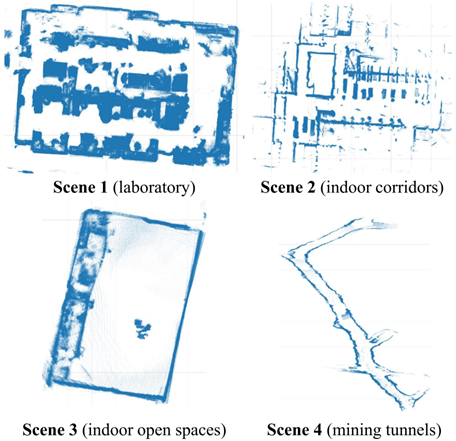}
  \caption{LiDAR point cloud global maps for 4 different scenarios \cite{ref31} }
  \label{fig:fig1}
\end{figure}

\subsection{Ground truth label generation}
Manual labeling often incurs significant time and labor costs, with low accuracy and poor consistency. To address this issue, Orr et al. and Jin et al. \cite{ref27,ref28} manually labeled a small portion of their datasets and trained an additional segmentation model to generate the remaining annotations.. While this approach reduces some manual effort, the training and fine-tuning of the segmentation model introduce additional time costs and potential instability. In this study, a label generation method based on occupancy grid maps is proposed to balance label quality and efficiency.

Inspired by Jin et al. \cite{ref30}, in this work, the ground truth labels are defined as the unobstructed field of view (FoV) in polar coordinates under the bird-eye view (BEV). In practical applications, the unobstructed FoV represents the maximum navigable area. On one hand, LiDAR point clouds can more accurately depict the 3D spatial structure of obstacles in the BEV. On the other hand, occlusion relationships are better represented in polar coordinates. The proposed ground truth label generation method leverages LiDAR point clouds and pose information over time. The pose information is directly provided by the Coloradar dataset, but alternative methods such as such as the iterative closest point (ICP) algorithm for point cloud registration or the approach proposed by Hess et al. \cite{ref32}, which utilizes LiDAR point clouds, IMU data, and loop closure constraints for globally optimized pose graph generation, can also be used. The ground truth label generation process is illustrated in Fig. 2 and involves the following steps:

\begin{figure} 
    \centering
    \includegraphics{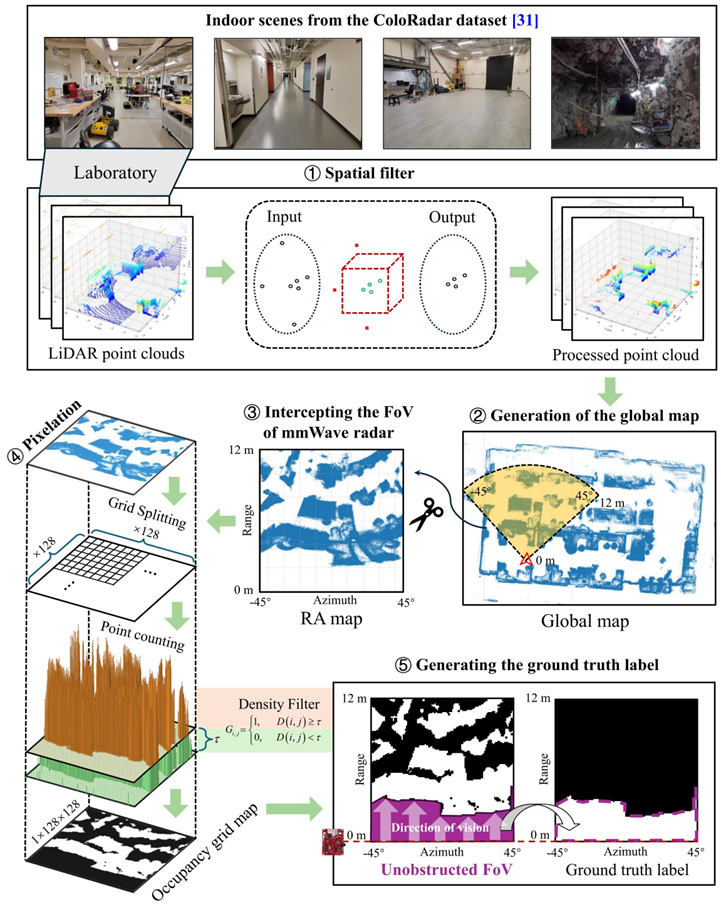}
  \caption{The generation process of the ground truth labels}
  \label{fig:fig2}
\end{figure}

\textbf{Constructing the Global Point Cloud Map: }Single-frame LiDAR point clouds are sparse and unevenly distributed across different distances. Constructing a global map integrates point clouds from multiple frames, clarifying distant spatial structures and increasing overall point cloud density. To enhance label generation efficiency, LiDAR point clouds are sampled every 2–10 frames for global map construction. Initially, point clouds are filtered based on intensity and spatial position, removing ground and ceiling points. The filtered point clouds are transformed into the global coordinate system using a transformation matrix $T$. Each point in $P_{\text{filtered}} = \left\{ p_{1}, p_{2}, \ldots, p_{n} \right\}
$
can be noted as $p_{i} = \left[ x_{i}, y_{i}, z_{i}, I_{i} \right]
$, then the point cloud set in the global coordinate system can be expressed as:

\begin{equation}
P_{\text{global}} = \bigcup_{k=1}^{m} \left\{ T_{k} \cdot p_{i}^{T} \mid p_{i} \in P_{\text{filtered}} \right\}
\end{equation}
\begin{equation}
T_{k} = \begin{bmatrix}
R_{k} & t_{k} \\
0 & 1 \\
\end{bmatrix}
\end{equation}

where \textit{m} is the number of the frames. $T_{k}
$, and $R_{k}$ are the transformation matrix, rotation matrix, and translation vector for the \textit{k}-th frame, respectively.

\textbf{Generation of range-azimuth map (RAM):} After obtaining the global map, height and intensity information are removed, and the point cloud is projected onto a 2D plane. At each moment, the coordinate origin is shifted to the sensor position, and the sensor orientation is aligned with the vertical axis. The point cloud is then transformed into polar coordinates and plotted in the RAM, with a distance range of 0–12 m and an azimuth angle range of -45° to 45°, representing the effective radar field of view (FoV).

\textbf{Conversion to Occupancy Grid Map: }To facilitate ground truth label generation at the pixel level, the RAM is converted into a 128×128 occupancy grid map, where each cell represents a spatial region in the RAM. Each grid cell has two states: 0 (free) and 1 (occupied). The occupancy grid map can be represented as a matrix  $G$  where $G \in \left\{ 0, 1 \right\}^{128 \times 128}
$. Each element 
$G_{i,j}
$ represents the label value at row \textit{i} and column \textit{j}, which is determined by following equation:

\begin{equation}
G_{i,j} = \begin{cases} 
1, & D_{i,j} \ge \tau \\
0, & D_{i,j} < \tau \\
\end{cases}
\end{equation}

Here,    denotes the number of points in the corresponding spatial region of    in the RAM, and $G_{i,j}
$ is the threshold depending on the scenarios. This process sets grid cells corresponding to obstacles to 1 and free spaces to 0, while reducing scatter in the spatio-temporal dimension. An erosion operation is performed to eliminate any spatial outliers.

\textbf{Ground truth label Generation:} The ground truth label    contains two categories, where 1 represents the unobstructed FoV, and 0 represents areas obstructed by obstacles. Here, the unobstructed FoV is defined as the region in each azimuth direction where no obstacles block the line of sight.

\section{Proposed Method}

\begin{figure} 
    \centering
    \includegraphics{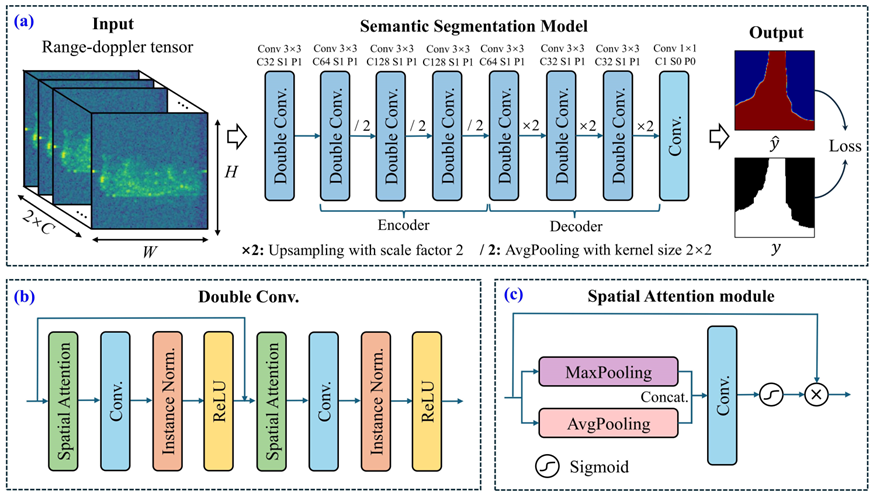}
  \caption{Architecture of the semantic segmentation mode}
  \label{fig:fig3}
\end{figure}

The input to our model is either the raw ADC data from the mmWave radar or the processed RD tensor and RA tensor. The real and imaginary parts of the data are separated and input simultaneously. Therefore, the tensor shape is (2×C)×H×W. The output of the model is a prediction matrix with shape H×W, where the height and width correspond to the distance range of 0-12 m and the azimuth range of -45° to 45° in polar coordinates under BEV. Each element of the prediction matrix represents the probability that the corresponding region belongs to the unobstructed FoV.
The model proposed in this work is based on U-Net, a widely used semantic segmentation architecture \cite{ref33}. As shown in Fig. 3, the encoder-decoder structure includes three downsampling modules in the encoder and three upsampling modules in the decoder. The main components of the model are double convolutional modules. Each double convolutional module consists of a 2D convolutional layer with a kernel size of 3×3 and padding of 1, followed by an instance normalization layer, a spatial attention module, and a ReLU activation function. 

In the downsampling modules, input feature maps are first reduced to half their original size using an average pooling layer, and then processed by a double convolutional module, doubling the number of channels. In the upsampling modules, input feature maps are enlarged to twice their original size using transpose convolution, followed by a double convolutional module that halves the number of channels.

Inspired by the Convolutional Block Attention Module (CBAM) \cite{ref34}, as illustrated in Fig. 3(c), the spatial attention module first applies global average pooling and global max pooling along the channel dimension of the input feature map, generating two single-channel feature maps. Max pooling captures the peak values across channels, while average pooling suppresses these peaks, highlighting background noise. These two pooled feature maps are then concatenated along the channel dimension to form a fused feature map, which is then processed by a 2D convolutional layer to generate the spatial attention map. The spatial attention map is converted into weights via a sigmoid layer and applied to the original input feature map. By incorporating the spatial attention module into the upsampling and downsampling blocks, the ability of the model to focus on relevant features at different scales of the RD tensor during forward propagation is enhanced, while attention to irrelevant features is reduced.

\section{Training Details}

Based on the waveform parameters of the mmWave radar used in the ColoRadar dataset, the raw ADC data can be converted into a 12×128×128 tensor, where the three dimensions correspond to the virtual channel, the chirp, and the sampling point dimension, respectively. The RD tensor is obtained from the raw ADC data by performing FFT sequentially along the sampling point and the chirp dimension. To fully utilize the information in the RD tensor and facilitate subsequent computations, the real and imaginary parts are separated and used as inputs. Ground truth labels are generated using the method described in Section 2.2. Additionally, all labels were manually inspected, with those of low quality or erroneous labels, along with their corresponding data, being manually removed. A few labels affected by noise were manually corrected. The size of each label is 128×128, represented as a matrix.

In this work, the segmentation task for the unobstructed FoV is a typical binary classification task for 2D images. The model is trained using a backpropagation algorithm, and the segmentation loss is defined as a weighted sum of binary cross-entropy (BCE) loss and dice loss \cite{ref35},which is a commonly used approach \cite{ref27,ref28,ref30}. The formula is as follows:

\begin{equation}
\mathcal{L}_{\text{seg}} = \lambda_{\text{bce}} \mathcal{L}_{\text{bce}} + \lambda_{\text{dice}} \mathcal{L}_{\text{dice}}
\end{equation}

where $\mathcal{L}_{\text{bce}}$ is the binary cross-entropy loss, and $\mathcal{L}_{\text{dice}}$ is the dice loss. Assume that for the $i$ th sample, the predicted value is$\hat{y}_{i}$, the label is $y_{i}$, and the total number of samples is $N$, the losses are defined as follows:

\begin{equation}
\mathcal{L}_{\text{bce}} = -\frac{1}{N} \sum_{i=1}^{N} \left[ y_{i} \cdot \log \left( \hat{y}_{i} \right) + \left( 1 - y_{i} \right) \log \left( 1 - \hat{y}_{i} \right) \right]
\end{equation}

\begin{equation}
\mathcal{L}_{\text{dice}} = 1 - \sum_{i=1}^{N} \frac{2 y_{i} \cdot \hat{y}_{i}}{y_{i}^{2} + \hat{y}_{i}^{2}}
\end{equation}

Herein, $\lambda_{\text{bce}}$ and $\lambda_{\text{dice}}
$ are the weight coefficients for the BCE and dice losses, respectively. In this work, $\lambda_{\text{bce}}$ and $\lambda_{\text{dice}}$ are set to 0.5 and 0.5, respectively.

The model is trained using the PyTorch framework on a workstation equipped with an Intel Xeon Silver 4210R CPU and an Nvidia GeForce RTX 2080Ti GPU. We used the Adam optimizer \cite{ref36} is employed, with an initial learning rate of 0.000314, which decreases gradually throughout the training process. The batch size is set to 16, and training is conducted for approximately 50 epochs.

\section{Experiment and Evaluations}
\subsection{Comparison with conventional point cloud-based method}

Figure 4 compares the environment representation between the conventional point cloud-based method and the proposed cross-modal semantic segmentation method. Here, the point cloud generation method chosen is the most commonly used conventional approach based on FFT and CFAR \cite{ref37}. Specifically, the ADC data cube undergoes FFT successively along the fast time, slow time, and virtual channel dimensions to obtain the range-Doppler velocity-angle heat map, followed by CFAR to detect peaks representing the detected targets, i.e., the spatial point cloud \cite{ref31}. This method is widely used due to its low computational overhead and ease of integration into hardware.

As shown in Fig. 4, the LiDAR point cloud is regarded as the ground truth and serves as the reference. The mmWave radar point cloud generated using the aforementioned method and the segmentation results of the proposed method for the unobstructed FoV are presented together to provide an intuitive comparison. All three outputs—LiDAR point cloud, mmWave radar point cloud, and segmentation results—are displayed in the same range-azimuth coordinate system. In the rightmost column, the segmentation results are superimposed onto the LiDAR point cloud to clearly demonstrate the performance of the proposed method in segmenting the unobstructed FoV. 

Essentially, point cloud-based environment perception methods detect and estimate the positions of reflective sources from echo signals through algorithms, presenting them in the form of point clouds. Formally, this is expressed as the detection of obstacles. In contrast, the environment perception method proposed in this paper attempts to use convolutional neural networks to extract free space within the FoV from the raw echo signals and reconstruct the contours of obstacles (segmentation boundaries) to some extent. For urgent applications like firefighting and rescue, segmenting free space offers a significant advantage over merely detecting obstacles, as it directly indicates navigable areas, eliminating the need for secondary processing to determine a path after obstacle detection.

Obviously, the environment perception based on conventional point clouds can outline the general structure of the environment to some extent; they tend to produce sparse and less accurate point clouds.. In complex indoor environments, noise points and multipath effects further degrade the quality of the point cloud. In contrast, the semantic segmentation-based method provides a more accurate and intuitive representation of the environment.

\begin{figure} 
    \centering
    \includegraphics{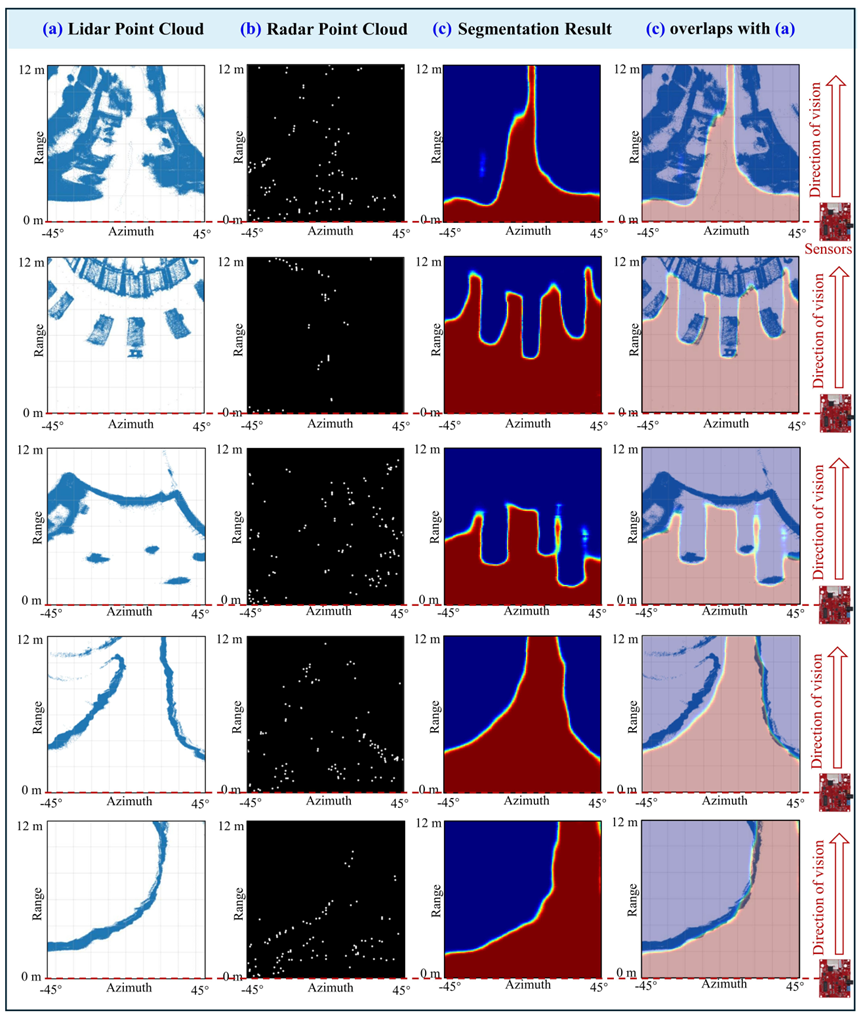}
  \caption{Comparison between conventional point cloud based and semantic segmentation methods}
  \label{fig:fig4}
\end{figure}

\subsection{Ablation Analysis}

At present, there is no existing segmentation model that is fully tailored to the segmentation tasks involved in the scenarios discussed in this paper. Fortunately, some models based on RD tensors have been proposed for classification \cite{ref29} or segmentation tasks \cite{ref28,ref38,ref39}. Although these works employed completely different MIMO methods or are applied to different tasks, certain modules from these models could be adapted for the proposed task, as they have been shown to enhance the understanding of abstract RD tensors. Therefore, this section compares the segmentation performance of the channel attention mechanism (CAM) module used by Orr \cite{ref27} and Jin et al. \cite{ref28}, the Swin transformer module used by Giroux et al. \cite{ref29}, and the spatial attention module used in the model proposed in this paper (see Fig. 3(c)).

In the controlled experiments, the proposed model serves as the baseline. In the first experiment group, the CAM module replaces the spatial attention module in the proposed model. In the second group, two successive Swin transformer modules replace the double convolution modules in the encoder, using the patch merging module for downsampling. following the approach of  Giroux et al. \cite{ref29}, with the same hyperparameters. 

Metrics such as accuracy, precision, recall, intersection over union (IoU), false alarm rate (FAR), and F1 score are employed for evaluation, as defined by Eqs. (7)-(12). Here, TP stands for true positive, TN for true negative, FP for false positive, FN for false negative, and n is the total number of samples in the test set \cite{ref40}.

\begin{equation}
\text{accuracy} = \frac{1}{n} \sum_{i=1}^{n} \frac{\text{TP}_{i} + \text{TN}_{i}}{\text{TP}_{i} + \text{TN}_{i} + \text{FP}_{i} + \text{FN}_{i}}
\end{equation}

\begin{equation}
\text{precision} = \frac{1}{n} \sum_{i=1}^{n} \frac{\text{TP}_{i}}{\text{TP}_{i} + \text{FP}_{i}}
\end{equation}

\begin{equation}
\text{recall} = \frac{1}{n} \sum_{i=1}^{n} \frac{\text{TP}_{i}}{\text{TP}_{i} + \text{FN}_{i}}
\end{equation}

\begin{equation}
\text{IoU} = \frac{1}{n} \sum_{i=1}^{n} \frac{\text{TP}_{i}}{\text{TP}_{i} + \text{FP}_{i} + \text{FN}_{i}}
\end{equation}

\begin{equation}
\text{F}_{1} = \frac{1}{n} \sum_{i=1}^{n} \frac{2\text{TP}_{i}}{2\text{TP}_{i} + \text{FP}_{i} + \text{FN}_{i}}
\end{equation}

\begin{equation}
\text{FAR} = \frac{1}{n} \sum_{i=1}^{n} \frac{\text{FP}_{i}}{\text{FP}_{i} + \text{TN}_{i}}
\end{equation}

The comparison results are shown in Fig. 5. Before calculating the metrics, the predicted results are thresholded, with values greater than the threshold set to 1 and values less than the threshold set to 0. In Fig. 5(a)-(f), the vertical axis represents the corresponding metric values, while the horizontal axis represents the threshold. In the first group, the spatial attention module is used (the proposed method). In the second group, the CAM module replaces the spatial attention module, and in the third group, the Swin transformer module replaces the dual convolution module in the proposed model, with patch embedding used for downsampling.

Overall, the model using the spatial attention module (first group) outperforms the model using the CAM module by approximately 2$\%$ across various metrics (second group). The model using the Swin transformer module (third group) is less sensitive to threshold variations, showing an advantage over the second group in certain threshold ranges. However, the first group consistently performs the best across all thresholds. IoU, a key performance metric for semantic segmentation, shows the most significant improvement, with the proposed model outperforming the others by about 3$\%$ (see Fig. 5(e)). Clearly, the spatial attention module is better suited for the task of unobstructed field of view segmentation under BEV compared to the CAM module and Swin transformer modules.

\begin{figure} 
    \centering
    \includegraphics{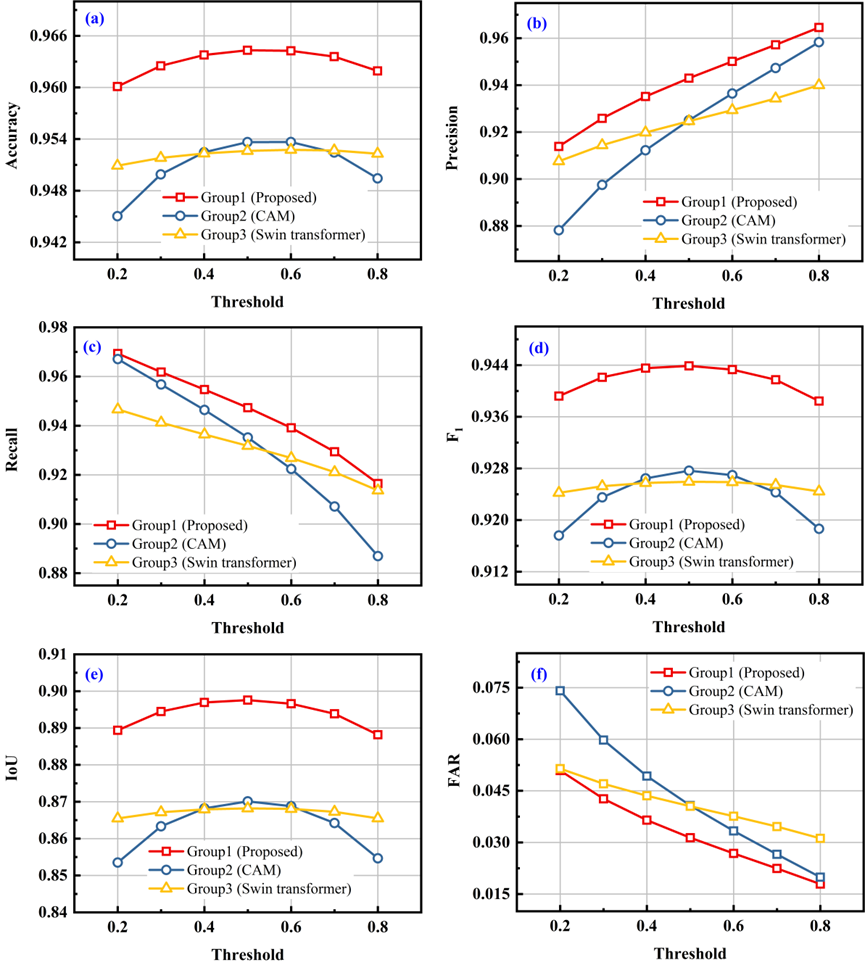}
  \caption{Comparative analysis of the proposed model (baseline) against models with CAM module and Swin transformer block across various performance metrics: (a) accuracy, (b) precision, (c) recall, (d) F1 score, (e) IoU and (f) FAR}
  \label{fig:fig5}
\end{figure}

To ensure a fair comparison, the number of learnable parameters and inference time are also considered. Fewer parameters and shorter inference times are crucial for mobile deployment and real-time inference. According to Fig. 6, the number of learnable parameters in the first group is very close to that of the second group, but the second group has longer inference times. Specifically, GPU inference time is 54$\%$ longer and CPU inference time is 43$\%$ longer. The third group has the largest number of learnable parameters, resulting in the longest inference time, which is undesirable for real-time applications. Overall, the proposed model (first group) not only has fewer parameters and a simpler structure but also achieves faster inference speeds, while delivering the best performance in this task This advantage is primarily attributed to the simplicity of the computational process of the spatial attention module, which achieves better results by cleverly utilizing only pooling layers and convolutional layers. Although the inference time here should be considered as a reference, since device performance is limited in practical applications, the proposed method still shows potential for real-time applications. These experiments were conducted in a consistent hardware environment using the same dataset, ensuring comparability.

\begin{figure} 
    \centering
    \includegraphics{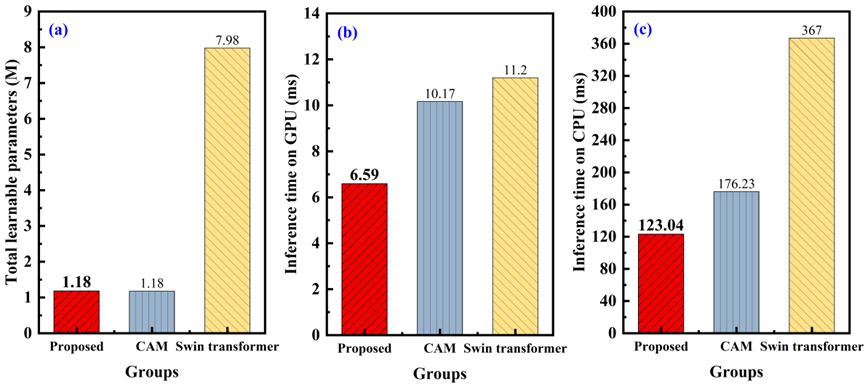}
  \caption{Comparison of the total parameters and inference time}
  \label{fig:fig6}
\end{figure}

To further illustrate the performance differences, Fig. 7 presents some specific prediction results alongside the Ground Truth images. The bright purple boxes in the figure highlight important details in the inference results. The color bar represents predicted values from 0 to 1, indicating the probability that the pixel belongs to the unobstructed FoV. According to Fig. 7, the proposed model exhibits more accurate segmentation performance, regardless of whether the environmental structure is complex (groups (a), (c), (d), (f), and (i)) or simpler (groups (b), (e), (g), (h), and (j)). In the second group, prediction errors are often large and patch-like, whereas in the third group, prediction errors are pixel-level and more scattered. This reflects the preference of the model for features at different scales during learning. For example, in groups (a) and (j), the second group captures large-scale features well but struggles with mid-scale features. while the third group favors small-scale details but has difficulties with mid-scale learning. In fact, mid-scale features, built on large-scale ones, are the most crucial for this task, while small-scale pixel-level details are mostly noise introduced during the labeling process. Comparatively, the prediction results of the first group are closest to the Ground Truth. Nonetheless, prediction errors are present and are unavoidable.

\begin{figure} 
    \centering
    \includegraphics{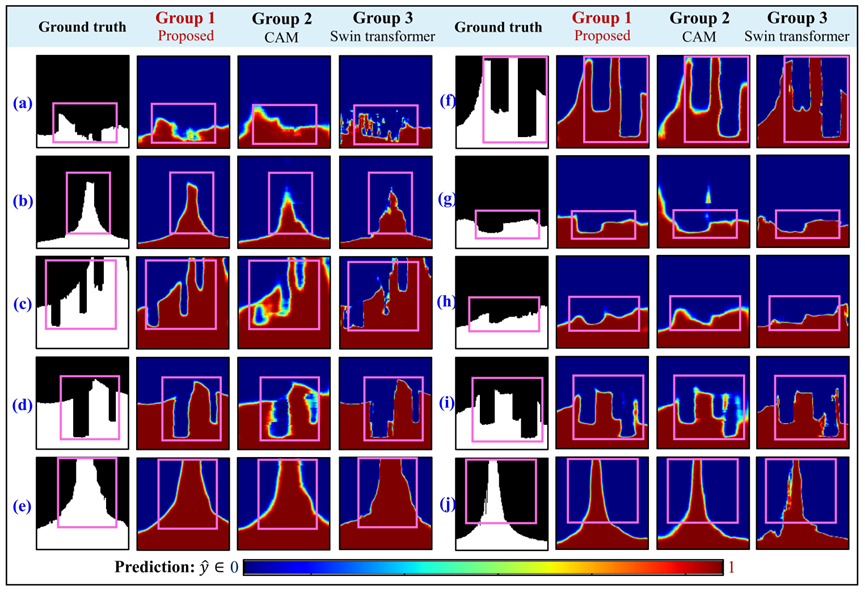}
  \caption{Demonstration of actual segmentation results}
  \label{fig:fig7}
\end{figure}

\subsection{Discussion on the characteristics of semantic segmentation models}

To further explore the characteristics of the semantic segmentation model in this task, the key metric, intersection over union (IoU), is discussed in terms of its variation with distance and angle. A known drawback of conventional signal point cloud generation methods based on FFT is that both angle estimation accuracy and angular resolution are affected by the azimuth angle. Eq. 13 describes the relationship between the phase difference between channels and the target azimuth angle. 

\begin{equation}
\Delta \varphi = \frac{2\pi d \sin(\theta)}{\lambda}
\end{equation}

Here, $\lambda$ is the wavelength of the carrier, $d$ is the spacing between virtual receiving antennas, $\Delta \varphi$ is the phase difference, and $\theta$ is the azimuth angle. It is important to note that the phase difference is not linearly related to the azimuth angle. The phase difference is most sensitive to changes in the azimuth angle when the azimuth angle is 0°, while it is least sensitive when the azimuth angle approaches 90°. Therefore, azimuth estimation becomes more error-prone at larger angles.

In contrast, the proposed signal processing method based on semantic segmentation is not particularly sensitive to changes in azimuth angle. According to Fig. 8, while the IoU metric varies slightly with angle, the variation is minor, with a difference of less than 2$\%$. The IoU is slightly lower at the sides compared to directly in front of the sensor. This effect is primarily influenced by the sample distribution in the dataset and will be further investigated in future work.

The variation of metric IoU with distance is shown in Fig. 9. It is evident that the IoU decreases as the distance increases. Specifically, the IoU attenuation range with distance is 27$\%$ for Group 3, 26$\%$ for Group 2, and 20$\%$ for the proposed model. Point cloud-based methods face similar issues, as shown in Fig. 4, where point cloud sparsity increases with distance due to a decrease in signal-to-noise ratio (SNR), resulting in fewer points at longer distances. Despite this, the proposed method shows certain advantages in environmental representation at longer distances.

\begin{figure} 
    \centering
    \includegraphics{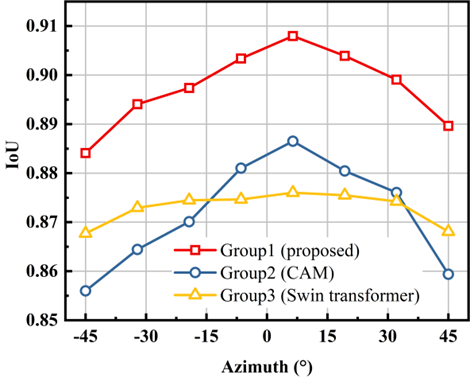}
  \caption{Variation of the metric IoU with azimuth}
  \label{fig:fig8}
\end{figure}

\begin{figure} 
    \centering
    \includegraphics{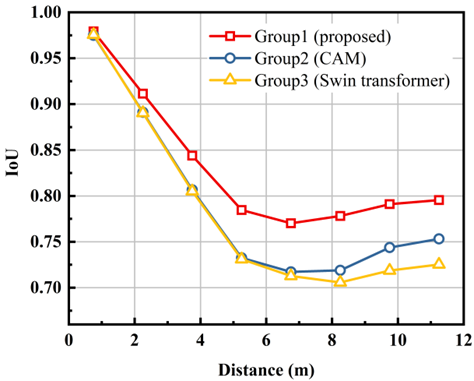}
  \caption{Variation of the metric IoU with distance}
  \label{fig:fig9}
\end{figure}

The consistent trend across models, where performance decreases with distance, is likely due to the dataset and the nature of the segmentation task itself. As the SNR of mmWave radar decreases with distance, segmentation becomes more challenging, particularly in mid-to-far regions where most of the segmentation details are located. Additionally, for the task of unobstructed FoV segmentation, the probability of occlusion is lower at close range, resulting in an uneven distribution of class 0 and class 1 in the ground truth labels. This imbalance will be addressed in future research.

The dataset used for training comprises four distinct scenarios: a laboratory, indoor corridors, indoor open spaces, and mining tunnels. In both the training and testing sets, the number of samples for each scenario is precisely equal. However, the segmentation performance of the models varies across these different settings, as detailed in Fig. 10. Specifically, Scene 1 corresponds to the laboratory, Scene 2 to indoor corridors, Scene 3 to indoor open spaces, and Scene 4 to mining tunnels.

\begin{figure} 
    \centering
    \includegraphics{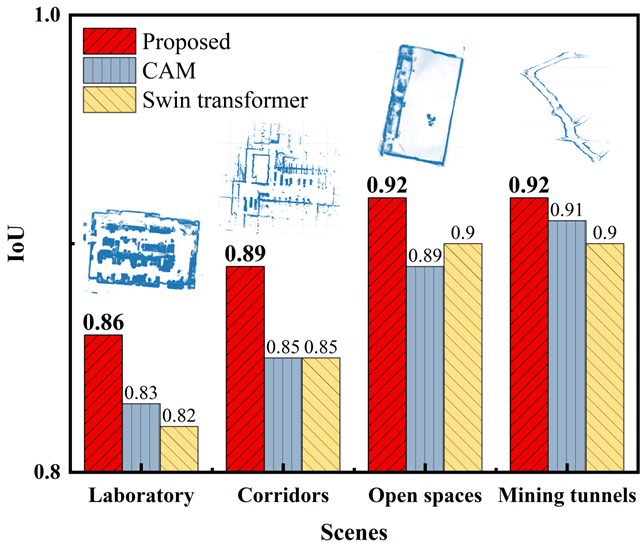}
  \caption{Comparison of the IoU metric in different scenes}
  \label{fig:fig10}
\end{figure}

According to Fig. 10 the segmentation performance of the various models is relatively similar in Scenes 3 and 4 and generally better than in Scenes 1 and 2. Specifically, Scene 3 is characterized by a large, rectangular indoor open space, while Scene 4 primarily consists of mining tunnels (see Fig. 1). Both of these scenes have simpler architectural structures and fewer obstacles, with walls mainly defining the unobstructed FoV. In contrast, Scenes 1 and 2 feature a wide variety of objects, including furniture of different shapes, sizes, and materials, which, along with architectural elements, define the boundaries of the unobstructed FoV.

Compared to simple elements like walls and floors, indoor objects vary greatly in type, geometry, size, and possess distinct reflective properties, which pose significant challenges to the semantic segmentation models. Despite these complexities, the proposed model demonstrates commendable segmentation performance across all scenes and surpasses other models on the IoU metric.

\subsection{Discussion on the proposed model with different input}

\begin{figure} 
    \centering
    \includegraphics{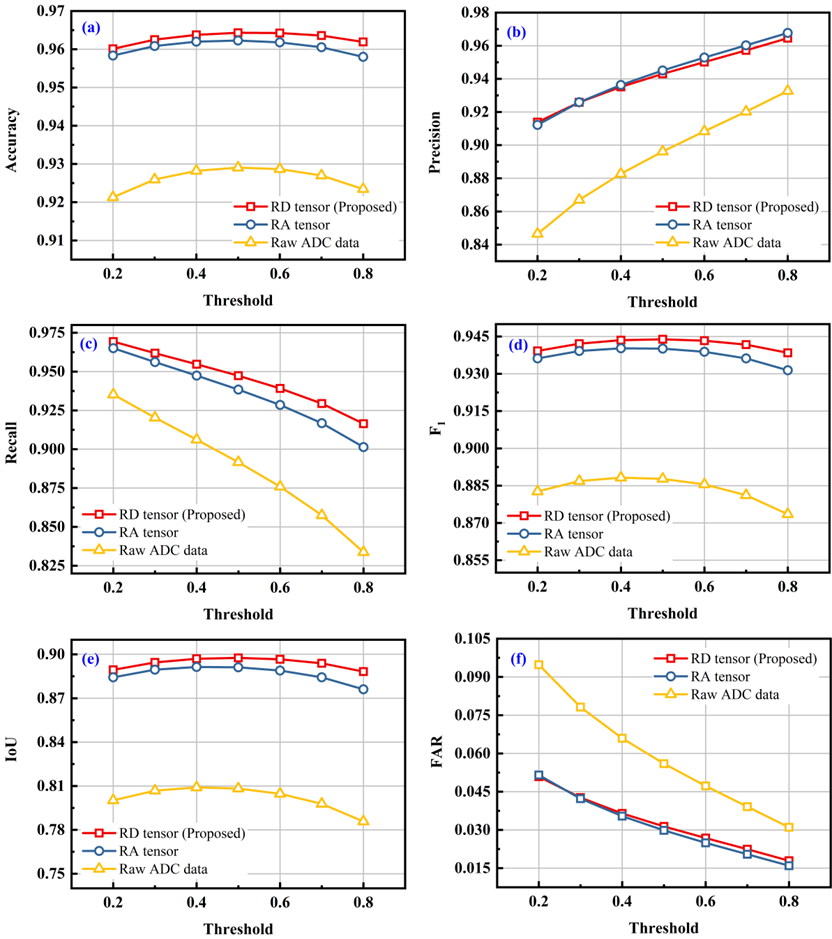}
  \caption{Comparison of the IoU metric in different scenes}
  \label{fig:fig11}
\end{figure}

In this section, the performance of the models was evaluated using ADC raw data and RA tensors as inputs, comparing them to the model trained with RD tensors as inputs. This analysis serves two purposes: first, to validate the feasibility of using other forms of mmWave radar data as inputs for the proposed model, and second, to demonstrate the superiority of using RD tensors as input. The metrics are presented in Fig. 11.

From Fig. 11, it is evident that using ADC raw data as the input of the model yields poor results, with all metrics significantly lower. This indicates that the model struggles to interpret and process the raw ADC data effectively. Using RA tensors as input is feasible, but the final model performance is slightly inferior to the model using RD tensors as input. Essentially, performing FFT on the ADC raw data across the fast time, slow time, and channel dimensions to obtain RD and RA tensors involves incorporating prior knowledge. The results shown in Fig. 11 suggest that the prior knowledge introduced during RD tensor conversion is beneficial and effective. However, the additional step of converting RD tensors to RA tensors via FFT along the channel dimension appears redundant. The FFT process can introduce errors such as spectral leakage, which, combined with over-processing, likely accounts for the observed performance degradation. Additionally, FFT operations add computational time, making RD tensors the optimal input choice for practical applications where performance and efficiency are crucial.

\section{Conclusion}

In this paper, a cross-modal semantic segmentation approach based on mmWave radar was employed for indoor environment perception in firefighting and rescue scenarios, focusing on segmenting the unobstructed FoV under BEV. Compared to conventional point clouds, the proposed method more intuitively and accurately captures and delineates indoor environments, with the segmentation performance being nearly unaffected by the azimuth. Additionally, a novel method for automatic label generation based on LiDAR point clouds is introduced, which significantly enhances labeling efficiency while ensuring label quality, and facilitating the collection and construction of large-scale datasets.
Given that existing neural network models are not well-suited for the segmentation tasks discussed in this paper, a new lightweight model is proposed. Modules previously employed to enhance the ability of the model to parse mmWave radar RD tensors, including the CAM module used by Orr et al \cite{ref27}. and the Swin transformer module used by Giroux \cite{ref29}., were compared with the proposed model. Results demonstrate that the proposed model achieves superior segmentation performance across most metrics. While segmentation performance decreases with increasing distance, the proposed model is the least affected by this factor.

Moreover, using raw ADC data as model input results in poor performance, whereas using RA tensors is feasible. However, the best segmentation results are achieved when RD tensors are used as input, avoiding the computational overhead and errors introduced by excessive preprocessing. In future work, smoke will be considered as a key factor, and a related dataset will be constructed and made publicly available. The impact of dataset structure on model performance will also be further investigated and mitigated. Additionally, current environment perception and segmentation approaches are limited to 2D. The development of motion estimation and SLAM algorithms based on cross-modal semantic segmentation remains a promising area for future exploration. These aspects will be elaborated on in subsequent research.


\end{document}